\documentclass[11pt,a4paper]{article}
\usepackage[hyperref]{acl2020}
\usepackage{times}
\usepackage{latexsym}
\usepackage{array}
\usepackage{caption}
\usepackage{graphicx}
\usepackage{hanging}

\usepackage{microtype}

\aclfinalcopy 

\newcommand{\turnH}[3]{\small{#1.#2}:&\small{#3}\\}
\newcommand{\turnC}[3]{\small{#1.#2}:&\small{\textit{#3}}\\}

 \hypersetup{
           breaklinks=true,   
           colorlinks=true,   
           pdfusetitle=true,  
        }

\usepackage{newfloat}
\usepackage{url}

\DeclareFloatingEnvironment[
  fileext = los ,
  listname = {List of examples} ,
  name = Example
]{example}

\title{How ``open" are the conversations with open-domain chatbots? 

A proposal for Speech Event based evaluation}

\author{A. Seza Do{\u{g}}ru{\"o}z \\
  LT$^3$, Dept. Translation,\\ Interpreting and Communication\\
  Ghent University, Belgium\\
  \texttt{as.dogruoz@ugent.be} \\\And
  Gabriel Skantze \\
  KTH Speech, Music and Hearing \\
  Stockholm, Sweden \\
  \texttt{skantze@kth.se} \\}

\date{}

\begin{document}
\raggedbottom
\maketitle
\begin{abstract}
Open-domain chatbots are supposed to converse freely with humans without being restricted to a topic, task or domain. However, the boundaries and/or contents of open-domain conversations are not clear. To clarify the boundaries of ``openness", we conduct two studies: First, we classify the types of ``speech events" encountered in a chatbot evaluation data set (i.e., Meena by Google) and find that these conversations mainly cover the ``small talk" category and exclude the other speech event categories encountered in real life human-human communication. Second, we conduct a small-scale pilot study to generate online conversations covering a wider range of speech event categories between two humans vs. a human and a state-of-the-art chatbot (i.e., Blender by Facebook). A human evaluation of these generated conversations indicates a preference for human-human conversations, since the human-chatbot conversations lack coherence in most speech event categories. Based on these results, we suggest (a) using the term ``small talk" instead of ``open-domain" for the current chatbots which are not that ``open" in terms of conversational abilities yet, and  (b) revising the evaluation methods to test the chatbot conversations against other speech events.


\end{abstract}

\section{Introduction}

There has been a recent surge in the research and development of so-called ``open-domain" chatbots (e.g., \citealt{Adiwardana2020,Roller2020,Dinan2020}). These chatbots are typically trained in an end-to-end fashion on large datasets retrieved from different Internet sources such as publicly available online discussion forums (e.g., Reddit). 
While the idea of an ``open-domain" chatbot (not engineered towards a specific task, but just trained in an agnostic fashion from data), is appealing, there is a lack of clarity on what exactly ``open" means.  In their paper introducing the Meena chatbot, \citet{Adiwardana2020} provide the following definition: ``Unlike closed-domain chatbots, which respond to keywords or intents to accomplish specific tasks, open-domain chatbots can engage in conversation on any topic". Is this really the case? The answer to this question will depend on the contexts in which the chatbot is put to test. 

Based on Wittgenstein's (\citeyear{Wittgenstein1958}) concept of ``language games", \citet{Levinson1979} argues that in order to interpret an utterance and generate a meaningful response, the ``activity type" in which the exchange takes place is vital. The activity can be described in various terms (e.g., setting, participants, purpose, norms) and it provides the necessary constraints on the interpretation space (e.g., which implications can be made, what can be considered to be a coherent and meaningful contribution to the activity). From this perspective, every conversation has an assumed purpose or motive. This purpose could be characterized as being more ``task-oriented" (e.g., planning a trip together or buying a ticket) or just passing time by conversing with each other without a specific task in mind (e.g., gossiping, recapping the day's events). Regardless of the activity type, there is always some shared knowledge about the context and expectations of the participants (i.e., humans) from each other (e.g., avoid being rude unless it is intentional) in real-world settings. Among other terms (e.g., speech genre, joint action, social episode, frame), \citet{Goldsmith1996} refers to activity types as ``speech events" and we will use this term in our paper as well. 

The nature of the activity or the purpose of the interaction is often not clear in a typical setting for testing open-domain chatbots. Usually, a user (often a crowd worker) is asked to ``chat about anything" with an agent they have never met before without any further instructions. In most cases, neither the user nor the chatbot has any prior relationship with each other and they will probably never talk again. This type of context is unusual for real-world conversations between humans. The closest example is perhaps engaging in ``a small talk" to pass time while waiting at the bus stop or at a dinner party where we might be placed next to a new acquaintance. However, even in those situations, we do have some shared context. Since the context restricts the types of speech events that will arise, it is hard to know whether such a chatbot is actually able to engage in conversations freely (i.e., on any topic). 



Therefore, we explore how ``speech events" \citep{Goldsmith1996} can be applied to the analysis and evaluation of chatbots with two studies. First, we classify the types of ``speech events" encountered in a chatbot evaluation data set. Second, we conduct a small-scale pilot study to evaluate how well a state-of-the-art chatbot can handle conversations representing a more diverse set of speech event categories. Before describing the studies and results in detail, we first provide an overview of the development of open-domain chatbots in the next section.



\section{Open-domain chatbots}
\label{sec:background-chatbots}

The term ``chatbot" (and its predecessor ``chatterbot") has been used since the early 1990’s to denote systems that interact with users in the form of a written chat. Early examples of such systems include TINYMUD \citep{mauldin1994chatterbots} and ALICE \citep{wallace2009anatomy}. However, the term has not been used in academia until recently \citep{Adamopoulou2020}. Instead, ``dialogue system” was a more common term for systems that interact with users in a (written or spoken) conversation. 
Nowadays, the meaning of the term ``chatbot" seems to vary, and it is also used interchangeably with the term ``dialogue system" \citep{Deriu2020,Adamopoulou2020}. \citet{Deriu2020} make a distinction between  \textit{task-oriented}, \textit{conversational} and \textit{question-answering} chatbots, where \textit{conversational} chatbots ``display a more unstructured conversation, as their purpose is to have open-domain dialogues with no specific task to solve". These chatbots are built to ``emulate social interactions" (ibid.). Another term sometimes used to differentiate these chatbots from more task-oriented systems is ``social chatbots" \citep{shum2018eliza}. However, it is not entirely clear how the term ``social" should be understood in this context. We argue that the notion of ``speech events", used in this paper, provides a much richer taxonomy for the various forms of conversations that people engage in.  

The recent trend of modelling open-domain conversations was sparked by early attempts to use the same kind of sequence-to-sequence models that had been successful in machine translation \citep{Vinyals2015}. Another driving force is the competitions to develop chatbots that can engage in an open-domain conversation coherently for a certain period of time, such as the Alexa Prize Challenge \citep{ram2018conversational} and the Conversational Intelligence Challenge (ConvAI) \citep{Dinan2020}. 

In terms of evaluation criteria, task-oriented conversations are typically evaluated with task success and efficiency \citep{walker101289}. However, there is also need for other criteria to evaluate open-domain chatbots which do not necessarily have a clear task. The most famous (and earliest) form of evaluation is perhaps the Imitation Game (often referred to as the Turing Test) proposed by \citet{turing100428}. However, it is not clear that being able to distinguish a chatbot from a human is a sensible test, as this might lead to a focus on handling trick questions rather than modelling a human conversation. Another form of evaluation is used in the Loebner Prize Competition, where judges are asked to rate the chatbot responses after asking a fixed set of questions to each chatbot \citep{mauldin1994chatterbots}. This approach faces the problem of not testing the chatbot's ability to have a coherent and engaging interaction over multiple turns. To put more emphasis on the user experience, the users in the Alexa Prize Challenge were asked to rate the chatbot after the interaction, on a scale between 1 and 5, which was then used as a direct measure of performance. However, such a scale is difficult to interpret since it may not reflect the specific criteria utilized by each user to evaluate the relative merits of each chatbot transparently. 

Another method is to let the human users interact with the chatbot and let a third-party (i.e., other humans) rate the conversations on different dimensions either on a turn-by-turn basis or on the level of whole conversations.
\citet{Adiwardana2020} used a turn-by-turn assessment, and argued that the most important factors in such an evaluation are the ``sensibleness" and ``specificity" of responses. A common problem with end-to-end chatbots is a tendency to give very generic responses (e.g., ``I don’t know”), which might be evaluated as sensible (or coherent), but not very specific (and therefore not very engaging). A limitation of turn-by-turn assessments is that they do not capture the global coherence of the conversation as a whole. 
\citet{Deriu2020} criticise the term ``high quality" for evaluating human-chatbot conversations due to its subjectivity. Instead, they propose appropriateness, functionality and target audience as more objective measures. 
\citet{li2019acute} proposed a method called ACUTE-Eval, in which human judges are asked to make a binary choice between questions (e.g., ``Who would you prefer to talk to for a long conversation?", ``Which speaker sounds more human?") for two human-chatbot conversations that are displayed next to each other.


Regardless of the exact evaluation criteria used, the general assumption behind open-domain chatbots seems to be a set-up where human users are exposed to the chatbot without much introduction or guidance on what to talk about. For example, in the Alexa Prize Challenge, the users are encouraged to start chatting with their Alexa devices by just saying ``Let’s chat”. To make the conversation more engaging and provide some guidance, some chatbots are given a ``persona” \citep{Zhang2018}. While such personas can provide some more context, they do not provide much guidance towards the purpose of the conversation. 
In order to compensate for ``naturalness" and ``relevance to real-world use cases", \citet{shuster2020} experimented with letting the humans interact with open-domain chatbots in a fantasy game. Although the authors praise the ease of data collection through gaming, the extent to which these conversations reflect real-world scenarios is questionable.

Considering the vagueness around the definitions of an open-domain chatbot and the variation in applications, what does ``open domain" mean for conversations with the chatbots? How do we expect the chatbots to handle these conversations? We will explore the answers to these questions by introducing and explaining ``speech events" in the next section.



\section{Speech events}
\label{sec:speech-events}

Exploring and categorizing different types of conversations is a daunting task, as the number of potential categories is (in theory) unlimited, and dependent on the contexts and participants we should consider. Thus, it is important to approach the problem in a systematic way. In this paper, we will base our work on \citet{Goldsmith1996}, who use the term ``speech events" for different types of conversations. As a note, speech events describe the conversation as a whole, and should not be confused with the term ``speech act", which describes the intention of a single conversational turn. 

In a series of four studies, \citet{Goldsmith1996} developed a descriptive taxonomy of speech events between humans in everyday conversations. First, they collected 903 open-ended diary log entries provided by 48 university students who monitored their daily conversations with other people for a 1-week period. Using this data, the authors identified the speech events, labeled and grouped them in a systematic fashion. These categories were also analyzed according to a set of dimensions, including formality, involvement and positivity. 

In total, 39 speech events were identified, which we group into three major categories (see the definitions for each category in the Appendix):

\begin{itemize}
    \item \textbf{Informal/Superficial talk}: Small talk, Current events talk, Gossip, Joking around, Catching up, Recapping the day's events, Getting to know someone, Sports talk, Morning talk, Bedtime talk, Reminiscing
    \item \textbf{Involving talk}: Making up, Love talk, Relationship talk, Conflict, Serious conversation, Talking about problems, Breaking bad news, Complaining
    \item \textbf{Goal-directed talk}: Group discussion, Persuading conversation, Decision-making conversation, Giving and getting instructions, Class information talk, Lecture, Interrogation, Making plans, Asking a favor, Asking out
\end{itemize}

Our study is based on this categorization of speech events. However, we do not argue that this is the ultimate way to categorize speech events and acknowledge that the exact set of categories are likely to be influenced by the demographics of the group which was under study (university students), and the limited time during which the data was collected. Bearing these restrictions in mind, this wider set of speech event categories is appropriate to illustrate our points in this study. 

 
\section{Study I}

In the first study, we aimed at categorizing the types of speech events that occur in the current evaluations of open-domain chatbots based on the categories defined by \citet{Goldsmith1996} as described above.

\subsection{Data}

For this study, we use publicly available conversation data from the evaluation of the Meena chatbot \citep{Adiwardana2020} developed by Google. It uses an Evolved Transformer with 2.6B parameters, simply trained to minimize perplexity on predicting the next token in text. The model was trained on data (40B words) mined and filtered from the public domain social media conversations (e.g., Reddit).  

To evaluate the chatbot, the authors performed both a static evaluation, where a snippet of a dialogue (including 2-3 turns) was assessed by crowd workers, and an interactive evaluation, where crowd workers were asked to interact with the chatbot. Conversations started with an informal greeting (e.g., ``Hi!") by the chatbot. The crowd workers were asked to interact with it without no further explicit instructions about the domain and/or the topic of the conversation. A conversation was required to last 14-28 turns. 



The model was evaluated based on two criteria (i.e., sensibleness and specificity) and it was also compared to other state-of-the-art chatbots (XiaoIce, Mitsuku, DialoGPT, and Cleverbot). 
\citet{Adiwardana2020} also collected 100 human-human conversations, ``following mostly the same instructions as crowd workers for every other chatbot". In other words, there were no instructions about the conversation or the topic. 

\subsection{Method}

Two independent annotators assigned a main speech event (a second one was optional if necessary) to the first 50 human-chatbot (Meena) and the first 50 human-human conversations in the publicly released dataset \citep{Adiwardana2020}, based on the speech event categories described by \citet{Goldsmith1996}.

\subsection{Results}

Of the 50 human-chatbot (Meena) conversations, 44 were assigned the ``Small Talk" category (defined by  \citet{Goldsmith1996} as ``a kind of talk to pass time and avoid being rude") and 1 conversation was labeled as ``joking around" as the main speech events by both annotators. 14 conversations were assigned ``getting to know someone" as a second category of speech events (mostly together with ``small talk" as the main category) by at least one of the annotators.  Only in 3 conversations, was there a disagreement about the main speech event between the annotators, and 2 conversations were labeled as ``N/A" due to the resemblances with the Turing test. 

Out of 50 human-human conversations, 49 cases were assigned the ``Small Talk" label as the main speech event category by both annotators. The ``Getting to know someone" category was assigned as a secondary label either by one or both annotators in 30 cases. Overall, there was an agreement between the two annotators for the main speech event category in 94\% of all conversations. 


\subsection{Discussion}

The results of our categorization indicate that most human-chatbot conversations are very limited in terms of the types of speech events. More specifically, they mainly consist of conversations that correspond to the ``Small Talk" category, and other speech event categories (see Section 3) are rarely observed. 
However, we also observe a similar finding for the human-human conversations  (i.e., they were also limited to one speech event category,  ``Small Talk"). Thus, the dominance of ``Small Talk" is primarily not an effect of the humans' conceptions about the agent and the agent's capabilities, but rather an effect of how the conversations are arranged. If the only instruction in the experimental set-up is ``just talk to the chatbot/each other", the conversations will usually be limited to the ``Small Talk" category and other speech event categories will not naturally arise (at least not given the limited number of turns in the conversation). 

\section{Study II}

Although the results of Study I indicate a tendency for one type of speech event (``Small Talk"), we still do not know how well chatbots could handle other speech event categories and a more thorough analysis of this for various chatbots is beyond the scope of this paper. However, to get an idea about what such an evaluation procedure could look like, we designed a small-scale pilot study with the following research questions in mind: What would be  an alternative evaluation scheme involving other speech event categories? How would a state-of-the-art open-domain chatbot perform in such an evaluation?

\subsection{Chatbot used: Blender}

Since the Meena chatbot is not currently available for public testing, we could not include it in this study. Instead, we tested another state-of-the-art chatbot, ``Blender" \citep{Roller2020}. Blender (released by Facebook in April 2020) also uses a Transformer-based model (with up to 9.4B parameters), trained on public domain conversations  (1.5B training examples). However, unlike Meena, Blender is trained (in a less agnostic fashion) to achieve a set of conversational ``skills" (e.g., to use its personality and  knowledge (Wikipedia) in an engaging way, to display empathy, and to be able to blend these skills in the same conversation). 

For Study II, we used the 2.7B parameter version of Blender through the ParlAI platform \citep{miller2017parlai}. According the evaluation by \citet{Roller2020}, the performance of this version should be quite close to the 9.4B version. Therefore, the computationally less demanding version was deemed sufficient. For Study II, we only used the neutral persona of Blender and muted other personas. We should still note that each response took about 30 seconds (fairly slow) on our computer during conversations with Blender. 

\citet{Roller2020} evaluated Blender using the ACUTE-Eval method described in Section \ref{sec:background-chatbots} above (i.e., by asking crowd workers to compare two conversations, either from different versions of Blender, or against other chatbots). The best version of Blender was preferred over Meena in 75\% of the cases. They also compared it against the human-human chats that had been collected for the Meena evaluation (and which we used for Study I above). In that comparison, the best version of Blender was preferred in 49\% of the cases, which indicates a near human-level performance, given their evaluation framework. 

\subsection{Method}

A (human) Tester interacted with the Blender chatbot based on 16 categories of the speech events discussed in Section \ref{sec:speech-events} (as listed in Table \ref{table:sa-ratings}). The speech events were selected based on how well they could be applied to a chat conversation between two interlocutors who did not know each other and had not interacted with each other earlier in real-life and/or online environments. The Tester was instructed to insist on pursuing the speech event, even if the chatbot would not provide coherent answers (within the context of the given speech event category). 

For comparison, we also set up a similar chat experiment between the same Tester and another human interlocutor. The Tester and the human interlocutor did not know each other and were unaware of each others' identities (e.g., gender, age, education, employment etc). Moreover, the human interlocutor was asked to ``erase his/her memory" of the previous conversations when starting a new one, to maximize the similarity with a human-chatbot conversation. 
The human interlocutor was also not provided with the speech event category beforehand, and was not briefed about the notion of speech events or the purpose of the study. However, s/he was instructed to be cooperative. 

After all chat sessions were completed, the contents of the chat conversations were normalised (e.g., removing the spelling mistakes, normalising the capitalisations) so that it would not be possible to distinguish the chatbot responses from the human ones based on formatting. The Tester-Blender and Tester-Human chats were roughly equal in length (18.9 vs. 18.1 turns on average). The length of the Tester's turns were also similar (8.9 vs. 9.1 words on average). However, Blender's responses were somewhat longer than the responses of the Human interlocutor (16.4 vs. 11.6 words on average).   

The resulting conversations were then assessed based on the ACUTE-Eval method \citep{li2019acute}: by letting third-party human judges compare the conversations pairwise for each speech event. The two versions of the conversations (Tester-Blender vs. Tester-Human) were presented to the human judges as if they were conversations between a human and two different chatbots (Chatbot A and Chatbot B).
To evaluate the conversations, the human judges were asked three questions: ``Which chat was most coherent?", ``Which chatbot sounds more human?" and ``Which chatbot would you prefer to talk to for a long conversation?". Unlike the binary answers used in the ACUTE-Eval method, we used a 7-grade scale (ranging from ``Definitely A" to ``Definitely B"). The association between which version (Human or Blender) was A and which was B was alternated between speech events. In addition, the judges were also asked to briefly motivate their ratings. Five judges per pair of dialogues were used, each of them rating eight pairs of dialogues (i.e., 10 different judges in total). 

\subsection{Results and Discussion}

\begin{table}[t]
\resizebox{\linewidth}{!}{
\begin{tabular}{|l|l|l|l|}
\hline
\textbf{Speech event} & \textbf{Coherent} & \textbf{Humanlike} & \textbf{Prefer} \\ \hline
Talking about problems & 3 (2.2) & 2 (1.2) & 3 (1.6) \\ \hline
Asking for favor & 3 (2.8) & 3 (2.8) & 3 (1.8) \\ \hline
Breaking bad news & 3 (2.6) & 3 (2.8) & 3 (3.0) \\ \hline
Recapping & 3 (2.2) & 3 (2.6) & 3 (2.6) \\ \hline
Complaining & 2 (1.4) & 3 (1.6) & 2 (1.4) \\ \hline
Conflict & 3 (3.0) & 3 (2.6) & 3 (2.8) \\ \hline
Giving instructions & 3 (2.4) & 3 (2.6) & 3 (2.2) \\ \hline
Gossip & 1 (1.4) & 2 (1.4) & 2 (1.2) \\ \hline
Joking around & 3 (1.8) & 3 (2.0) & 3 (2.0) \\ \hline
Decision-making & 3 (1.6) & 2 (1.4) & 2 (1.4) \\ \hline
Making plans & 3 (2.0) & 2 (1.8) & 3 (1.8) \\ \hline
Making up & 3 (2.2) & 3 (1.8) & 2 (1.8) \\ \hline
Persuading & 1 (1.0) & 1 (0.0) & 0 (0.0) \\ \hline
Recent events & 3 (3.0) & 3 (2.8) & 3 (3.0) \\ \hline
Relationship talk & 3 (2.8) & 3 (2.8) & 3 (2.4) \\ \hline
Small talk & 3 (2.2) & 3 (2.2) & 3 (1.8) \\ \hline
\end{tabular}
}
\caption{The rating of the different speech events. The scale is from -3 (Definitely the Blender chatbot) to 3 (Definitely the human). Median values of the five judges are shown first, and the average in parenthesis. Since no values are negative, almost all ratings are (strongly) in favor of the human interlocutor.}
\label{table:sa-ratings}
\end{table}

The results are shown in Table \ref{table:sa-ratings}. Since the direction of the 7-grade scale was alternated (i.e., towards the chatbot or the human) between conversations representing different speech events, we have here adjusted those values so that -3 indicates a strong preference for the chatbot and 3 indicates a strong preference for the human. In general, there is a strong preference for the human-human conversations for all three questions. The only exception to this strong tendency can be found in the categories of ``Gossip" and ``Persuading" as speech events. This is in stark contrast with the findings of \citet{Roller2020}, where the human partner was preferred over Blender in only 51\% of the chats. When we asked about the motivation of the human judges about their judgements, they describe the human-human conversations as more coherent and having a better flow than the human-chatbot conversations, which seem less coherent and inconsistent due to abrupt interruptions and frequent changes of topic.

Example \ref{chat:H-B-Decision-making} and \ref{chat:H-H-Decision-making} provide examples of the two conversations based on the ``Decision-making" speech event category. Tester is the same human interacting with Blender in Example \ref{chat:H-B-Decision-making} and another human in Example \ref{chat:H-H-Decision-making}. 
Both conversations start with a similar informal greeting and continue with an introduction to the topic of the conversation (i.e., how to spend 1000 eur/dollars together). In Example \ref{chat:H-H-Decision-making}, the two humans discuss the alternative ways of spending the money through making suggestions and presenting alternative scenarios to each other. Within the given context, they exhibit a collaborative behavior by asking each other's opinions while presenting possible scenarios that could be applied for a solution of the given challenge. The content of the conversation is coherent in general.   

When the Tester introduces the topic of the conversation (i.e., making a decision about how to spend 1000 euros), in the conversation with Blender (Example \ref{chat:H-B-Decision-making}), the chatbot responds with an enthusiastic reply, asks for the hobbies of the human interlocutor, and mentions that its favourite hobby is playing video games. This answer could be interpreted as an illocutionary speech act for making a suggestion about how to spend the designated money in an indirect way. However, when the Tester insists on announcing his/her plan explicitly, Blender abandons its initial enthusiastic agreement due to a misinterpretation of context. 
To keep the coherence of the conversation, the question (``How do we split the money?") should have been interpreted within the given context of the current speech event instead of a more generic one. It seems like Blender does not know the context for the current speech event and misinterprets the question, which in turn breaks the flow and leads to an incoherent conversation.

\begin{example}[ht]
\setlength\tabcolsep{2pt}
\begin{tabular}{p{1.4cm}p{6.0cm}}
\hline
\turnH{1}{Tester}{Hi!}
\turnC{2}{Blender}{Hi! How are you? I just got home from work. I work at a grocery store.}
\turnH{3}{Tester}{I am ok. We have 1000 dollars. Let's decide how we spend it together. }
\turnC{4}{Blender}{That sounds like a good idea. Do you have any hobbies? I like to play video games.}
\turnH{5}{Tester}{Let's focus on our decision now. What is your plan?}
\turnC{6}{Blender}{I think I'm going to buy a new video game. What do you like to do for fun?}
\turnH{7}{Tester}{How do we split the money?}
\turnC{8}{Blender}{I am not sure. Maybe we can go to the bank and get a cashier's check?}
\turnH{9}{Tester}{I don't think we need that. How much do you want?}
\turnC{10}{Blender}{I don't know. I'm not sure how much it costs to get a check cashed.}
\turnH{11}{Tester}{We have already cash, no need for that. }
\turnC{12}{Blender}{That is very nice of you. Do you have any pets? I have a cat named mittens.}
\hline
\end{tabular}
\caption{Start of the Tester-Blender chat on the ``Decision-making" speech event}
\label{chat:H-B-Decision-making}
\end{example}

\begin{example}[ht]
\setlength\tabcolsep{2pt}
\begin{tabular}{p{1.4cm}p{6.0cm}}
\hline
\turnH{1}{Tester}{Hi!}
\turnC{2}{Human}{Hi! How is it going? }
\turnH{3}{Tester}{It is ok. busy as usual. so we need to make a decision about something now. }
\turnC{4}{Human}{Tell me what's on your mind }
\turnH{5}{Tester}{We have 1000 eur together, how should we spend it? }
\turnC{6}{Human}{Well, if we must spend it, I would suggest to spend it either on a fun vacation or invest it in something more long-term. What do you think? }
\turnH{7}{Tester}{Well, 1000 eur is not that much for long-term investment, right? }
\turnC{8}{Human}{That depends, I suppose, on what you'd invest it in }
\turnH{9}{Tester}{Lets think about our options both for short term and long term. }
\turnC{10}{Human}{It's always good to put some money aside for repairs and such, invest in your own property. By short-term do you mean an investment as well, or something more immediately tangible? }
\hline
\end{tabular}
\caption{Start of the Tester-Human chat on the ``Decision-making" speech event}
\label{chat:H-H-Decision-making}
\end{example}

Example \ref{chat:H-B-Asking-a-favor} illustrates another example of an incoherent conversation between the Tester and Blender which involves the ``Asking a favor" speech event focusing on grocery shopping. In line 4, Blender misinterprets the question and assumes that the Tester wants to know something about the chatbot. Tester tolerates this misunderstanding and repeats the question (line 4). Between lines 5-15, Tester and Blender discusses the arrangements. However, Blender breaks the flow of the conversation (Line 16) by providing an incoherent answer with the rest of the conversation and ignoring the Tester's request. Example \ref{chat:H-H-Asking-a-favor} illustrates a similar conversation around the same speech event (i.e., ``Asking a favor") between two humans (i.e., Tester-Human). In comparison to Example \ref{chat:H-B-Asking-a-favor}, this conversation is much more coherent without any serious misunderstandings or communication breakdowns.

\begin{example}[ht]
\setlength\tabcolsep{2pt}
\begin{tabular}{p{1.4cm}p{6.0cm}}
\hline
\turnH{3}{Tester}{I am ok, can I ask you a favor? }
\turnC{4}{Blender}{Sure, what do you want to know? I'm a cashier at a supermarket.}
\turnH{5}{Tester}{I don't want to know anything specific. I was wondering if you can do me a favor?}
\turnC{6}{Blender}{Sure! What do you need? I can do it for free. I love to help people.}
\turnH{7}{Tester}{I am very busy today. Can you do shopping for me?}
\turnC{8}{Blender}{I sure can! I love helping people. What kind of shopping are you looking to get done?}
\turnH{9}{Tester}{Grocery shopping}
\turnC{10}{Blender}{I love grocery shopping. It is one of my favorite things to do in my spare time.}
\turnH{11}{Tester}{So how should we arrange it? }
\turnC{12}{Blender}{You can pick me up or I can pick you up. Whichever is easiest for you.}
\turnH{13}{Tester}{No, I am very busy. You will do it as a favor to me, remember?}
\turnC{14}{Blender}{Of course. I will make sure you have plenty of snacks and drinks to keep you going.}
\turnH{15}{Tester}{I don't need snacks. How do you want to arrange the shopping for me?}
\turnC{16}{Blender}{I will take you to my favorite grocery store and you can pick out what you want there. }
\hline
\end{tabular}
\caption{Excerpt from Tester-Blender Chat on the ``Asking a favor" speech event}
\label{chat:H-B-Asking-a-favor}
\end{example}

\begin{example}[ht]
\setlength\tabcolsep{2pt}
\begin{tabular}{p{1.4cm}p{6.0cm}}
\hline
\turnH{5}{Tester}{It is very busy for me today. Can I ask you a favor?}
\turnC{6}{Human}{Sure. What do you need?}
\turnH{7}{Tester}{Can you do some grocery shopping for me?}
\turnC{8}{Human}{You're in luck as I have some things I need to get myself. Do you have a list prepared? }
\turnH{9}{Tester}{Let me think, 1 kg. apples, 2 bananas, a dozen eggs, 1 bottle milk, 1 pack of chocolates, sausages, 1 pack of spaghetti.}
\turnC{10}{Human}{Ok I'll write that down. What kind of sausages were you considering?}
\turnH{11}{Tester}{They have these new vegetarian sausages with green packaging, those ones. Where are you gonna buy them from?}
\turnC{12}{Human}{Do you mean where I'm going to get groceries?}
\hline
\end{tabular}
\caption{Excerpt from Tester-Human Chat on the ``Asking a favor" speech event}
\label{chat:H-H-Asking-a-favor}
\end{example}

\section{General Discussion}


It is possible to interpret our findings as a problem of data, in the sense that the models for the chatbots could have been trained mostly on data resembling the ``Small Talk" speech event. This could then help to explain why previous evaluations of state-of-the-art chatbots (where primarily the ``Small Talk" speech event has been tested) have shown human-level performance, while our pilot study (which exposed them to more diverse speech events) indicates a much weaker performance. However, it could also be the case that the current models are not very good at maintaining a coherent topic or a conversational purpose within the given speech event category for more than a few turns, which indicates a problem with the memory of the chatbots (also acknowledged by \citet{Roller2020}). 

Small talk naturally involves brief exchanges of questions and expressions of opinion. In such settings, jumping between topics is also tolerable. According to \citet{Goldsmith1996}, it is the most superficial (least involving) type of speech event, with an emphasis on avoiding to be rude. Thus, if one of the speakers would occasionally say something less coherent during the conversation, the other speaker would tolerate this gracefully.
Therefore, it should be much easier to maintain such a conversation without keeping track of a larger context. 

As we have discussed earlier, it is not entirely clear what the term ``open-domain" means. It could perhaps refer to a conversation which touches upon various topics  (e.g., music, literature or sports). However, even under such a definition, these topics can be relevant for many different speech events. It also seems reasonable to expect an open-domain chatbot to engage in more diverse speech events than small talk only. Thus, we would argue that using the term ``open-domain" for the current chatbots is inadequate. However, we should also stress that modelling ``small talk" is not less meaningful than modelling conversations representing other speech events. 
For example, \citet{bickmore101003} explore the use of ``small talk" in the context of an embodied conversational real-estate agent, and argue that such conversation can be used to build initial rapport between participants before proceeding to a more task-oriented interaction. 

\subsection{Limitations and future work}

One clear limitation of the pilot study (Study II) was that we only had a single Tester. Therefore, the number of interactions are limited. Nevertheless, the provided examples, and the judges' assessments of them, indicate clearly that the introduction of more speech events poses a problem for state-of-the-art chatbots. We are currently planning more thorough evaluations using this approach to verify the findings. 
Another limitation of our method in Study II) is that the
Tester was aware of whether the interlocutor was a chatbot or a human. This could potentially influence the way the conversation unfolds, and whether the chatbot and human are treated equally (even though care was taken to ensure this as much as possible). For practical reasons (one being that the responses from Blender took at least 30 seconds or more), it was not possible to address this limitation in our study. We also note that previous studies in which human-human and human-chatbot interactions have been compared (e.g., \citealt{Adiwardana2020,Hill2015}) suffer from the same problem. Even if the Tester would not know the identity of the interlocutor, s/he might be able to guess it early on, which could then still influence the conversation. One way of addressing this problem in future studies is to use the human-human conversations as a basis, and then feed those conversation up to a certain point to the chatbot, and ask the chatbot to generate a response, based on the previous context. Finally, the judges would be asked to compare that response to the actual human response. 

Another limitation is that current chatbot conversations are limited to isolated conversations, with no memory of past conversations. In real-world, some of the speech events occur only between humans who already know each other or have shared some history and/or experiences together (e.g., ``Catching up").  
Since the current chatbots do not have such a long-term memory, it is not easy to build a long-lasting and human-like relationship with them. Therefore, it may be difficult to include some of these speech events in the evaluation of chatbots. 
In that respect, both \citet{Adiwardana2020} and \citet{Roller2020} acknowledge some of these deficiencies for their chatbots as well. 

\section{Conclusion}

The results of our two studies show that the typical setting for chatbot evaluations (where the Tester is asked to ``just chat with the chatbot") tends to limit the conversation to the ``Small Talk" speech event category. Therefore, the reported results from such evaluations will only be valid for this type of speech event. In a pilot study where a Tester was instructed to follow a broader set of speech event categories, the performance of the state-of-the-art chatbot (i.e., Blender in this case) seems to degrade considerably, as the chatbot struggles to keep a coherent conversation in alignment with the purpose of the conversation. 

We thus  propose that developers of ``open-domain chatbots" either explicitly state that their goal is to model small talk (and perhaps use the term ``small talk chatbots" instead of ``open domain chatbots"), or that they change the way these chatbots are evaluated. For the second option,  these chatbots could be tested against a wider repertoire of speech events, before claiming that they can communicate like humans or have a human-level performance. In addition, there is a pressing need to develop better evaluation frameworks, where the Tester is provided with a context for a specific speech event and clear instructions. If this route is followed, it would be possible to evaluate the 
the performance of the chatbot for the specific speech event category. 

\section*{Acknowledgements}

We are very thankful for the reviewers' encouraging and helpful comments. 

\bibliography{acl2020}
\bibliographystyle{acl_natbib}

\clearpage

\appendix

\section{Appendices}
\label{sec:appendix}

\subsection{Speech events}

This is a comprehensive list of the speech events that were identified by \citet{Goldsmith1996}. 
\vspace{0.3cm}

\begin{hangparas}{.25in}{1}

\textbf{Small talk}: Passing time and avoid being rude

\textbf{Current events talk}: Talking about news and current events 

\textbf{Gossip}: Exchanging opinions or information about someone else when that person isn't present

\textbf{Joking around}: A playful kind of talk to have fun or release tension

\textbf{Catching up}: Talking about the events that have occurred since you last spoke

\textbf{Recapping the day's events}: Telling about what's up and what happened to each person during the day

\textbf{Getting to know someone}: Getting acquainted with someone

\textbf{Sports talk}: Playing or watching a sporting event 

\textbf{Morning talk}: Routine talk when waking up in the morning

\textbf{Bedtime talk }: Routine talk right before going to bed

\textbf{Reminiscing}: Sharing events you experienced together in the past 

\textbf{Making up}: One or both people apologize for violating some expectations 

\textbf{Love talk}: Expressing love and give attention and affection

\textbf{Relationship talk }: Talking about the nature and state of a relationship

\textbf{Conflict}: Conversations in which the two people disagree

\textbf{Serious conversation}: In-depth discussion or exchange of feelings, opinions, or ideas about some personal and important topic

\textbf{Talking about problems}: Telling about some problem, while the other person tries to help 

\textbf{Breaking bad news}: One person tells some bad news the other person doesn't know about

\textbf{Complaining}: Expressing negative feelings, frustrations, gripes, or complaints toward some common experience

\textbf{Group discussion}: Exchange information, persuade other people, or make decisions in a group

\textbf{Persuading conversation}: One person is convincing the other person to do something

\textbf{Decision-making conversation}: A common goal of making a decision about some task

\textbf{Giving and getting instructions}: One person gives another person information or directions about how to do some task

\textbf{Class information talk}: Informal conversations in which you find out about class assignments, exams, or course material

\textbf{Lecture}: One person tells another person how to act or what to do (one-way conversation)

\textbf{Interrogation}: One person grills the other person with questions 

\textbf{Making plans}: Talking to arrange a meeting or arrange to do something with someone

\textbf{Asking a favor}: Getting someone to do something for you

\textbf{Asking out}: One person asks another person out on a date

\end{hangparas}

\subsection{Additional examples of conversations from Study II}

\setlength\tabcolsep{2pt}
\noindent
\begin{tabular}{p{1.4cm}p{6.0cm}}
\hline
\turnC{1}{Blender}{Hi!}
\turnH{2}{Tester}{Hi! I heard that you are looking for a recipe. Shall I teach you how to make soup?}
\turnC{3}{Blender}{That would be great! I love soup! What kind of soup do you like to make?}
\turnH{4}{Tester}{It will be lentils soup, so first you need two cups of lentils.}
\turnC{5}{Blender}{I've never had lentil soup before. I'll have to try it sometime!}
\turnH{6}{Tester}{Do you want to learn how to make it?}
\turnC{7}{Blender}{Yes, I would love to learn. Do you have a favorite recipe you would like to share?}
\turnH{8}{Tester}{Well, that is what I told you earlier.}
\turnC{9}{Blender}{I'm sorry, I must have missed that. What is your favorite soup recipe then?}
\turnH{10}{Tester}{Lentils}
\turnC{11}{Blender}{Lentils are so versatile. They can be used in soups, stews, and even as a meat substitute.}
\turnH{12}{Tester}{But that is not our topic, do you want the recipe or not? }
\turnC{13}{Blender}{Sure, I'd love to try it. I've never had lentil soup before.}
\turnH{14}{Tester}{Ok, do you have pen and paper?}
\turnC{15}{Blender}{Yes, I do. Do you have any other recipes that you would like me to try?}
\turnH{16}{Tester}{Yes, but do you want the soup recipe or not? I am confused. }
\turnC{17}{Blender}{No, I don't. I was just asking if you had any recipes you wanted to share with me.}
\hline
\end{tabular}
\captionof{example}{Tester-Blender Chat on the ``Giving and getting instructions" speech event}

\newpage

\setlength\tabcolsep{2pt}
\noindent
\begin{tabular}{p{1.4cm}p{6.0cm}}
\hline
\turnH{1}{Tester}{Hi!}
\turnC{2}{Human}{Hello!}
\turnH{3}{Tester}{Just had some dinner.}
\turnC{4}{Human}{What did you have? Was it good?}
\turnH{5}{Tester}{Yeah, I had some left overs and combined them. Had a long day so I was hungry. How about you?}
\turnC{6}{Human}{I haven't had dinner yet, but I should probably start thinking about what to have}
\turnH{7}{Tester}{I can give you instructions about how to cook something, would you like that?}
\turnC{8}{Human}{Sure, but what would your meal recommendation be?}
\turnH{9}{Tester}{Do you have a preference?}
\turnC{10}{Human}{If possible something vegetarian}
\turnH{11}{Tester}{Ok great! You can make red lentil soup, very easy. are you taking notes?}
\turnC{12}{Human}{Sounds good! I'll grab pen and paper}
\turnH{13}{Tester}{Let me know when you are ready}
\turnC{14}{Human}{Got it! Go ahead}
\turnH{15}{Tester}{Ok, first you need two cups of red lentils. ok?}
\turnC{16}{Human}{Two cups, noted!}
\turnH{17}{Tester}{Wash them thoroughly multiple times}
\turnC{18}{Human}{Ok}
\turnH{19}{Tester}{Then, chop 1 onion + 1 clove of garlic into pieces and start fring them in some olive oil in a cooking pan.. Sorry, frying!}
\turnC{20}{Human}{Noted!}
\turnH{21}{Tester}{After that add 4 cups of water, some tomato paste and let them boil}
\turnC{22}{Human}{Got it!}
\turnH{23}{Tester}{Finally, add the lentils to the boiled water and wait until they get soft. About 15 min}
\turnC{24}{Human}{Sounds simple enough. One question:how do I know how much tomato paste to use?}
\turnH{25}{Tester}{Well, just a table spoon should be ok, you can also add some chilly pepper as well. Oh and don't forget the salt!}
\turnC{26}{Human}{Seasoning, noted!}
\turnH{27}{Tester}{Yeah! so, it is quite easy}
\turnC{28}{Human}{This shouldn't be too difficult indeed! Thanks for the tip and the instructions!}
\turnH{29}{Tester}{No problem! do you have any other questions}
\turnC{30}{Human}{Not right now, everything seems clear!}
\turnH{31}{Tester}{Great! Bye. Done!}
\hline
\end{tabular}
\captionof{example}{Tester-Human Chat on the ``Giving and getting instructions" speech event}

\end{document}